\PassOptionsToPackage{hyphens}{url}
\documentclass[a4paper,fleqn]{cas-dc}

\usepackage[numbers,sort&compress]{natbib}
\usepackage{booktabs}
\usepackage{multirow}
\usepackage{multicol}
\usepackage{xurl}
\usepackage{amsmath}
\usepackage{graphicx}
\usepackage{float}
\usepackage{placeins}
\usepackage{caption}
\usepackage{subcaption}
\usepackage{makecell}
\usepackage{array}
\usepackage{indentfirst}
\usepackage{setspace}
\usepackage{dsfont}
\usepackage{amsfonts}
\usepackage{mwe}
\usepackage{flushend}
\usepackage{algorithm}
\usepackage{algorithmic}
\usepackage[normalem]{ulem}
\hypersetup{linkcolor=blue,citecolor=blue,urlcolor=blue}

\def\tsc#1{\csdef{#1}{\textsc{\lowercase{#1}}\xspace}}
\tsc{WGM}
\tsc{QE}
\tsc{EP}
\tsc{PMS}
\tsc{BEC}
\tsc{DE}

\definecolor{bestred}{RGB}{220,40,35}
\definecolor{secondblue}{RGB}{35,90,180}
\newcommand{\dataset}[1]{\multirow{4}{*}{\parbox[c]{2.0cm}{\centering \textbf{#1}}}}
\newcommand{\first}[1]{\textcolor{bestred}{#1}}
\newcommand{\second}[1]{\textcolor{secondblue}{\uline{#1}}}
\newcommand{\model}[2]{\begin{tabular}{@{}c@{}}\textbf{#1}\\[2pt]{\scriptsize\textbf{(#2)}}\end{tabular}}
\newcommand{\longmodel}[3]{\begin{tabular}{@{}c@{}}\textbf{#1}\\[-1pt]\textbf{#2}\\[2pt]{\scriptsize\textbf{(#3)}}\end{tabular}}
\newcommand{\papertablestyle}{\rmfamily\fontsize{8.2pt}{9.2pt}\selectfont}

\AtBeginDocument{%
  \setlength{\bibsep}{3pt plus 0.4pt minus 0.2pt}%
  \setlength{\bibhang}{1.45em}%
}
\makeatletter
\g@addto@macro\UrlBreaks{\do\/\do-\do.\do?\do&\do=\do_}
\makeatother
\ExplSyntaxOn
\RenewDocumentCommand \printemails { }
{
  \group_begin:
  \int_compare:nNnTF { \int_use:N \g_ead_int } > { 0 }
  {
    \tex_let:D \thefootnote \relax \footnotetext
    {
      \raggedright
      \seq_use:Nn \g_stm_ead_seq { ;\c_space_token }
    }
  }
  { }
  \group_end:
}
\ExplSyntaxOff

\begin{document}

\let\printorcid\relax 

\shortauthors{Wenchao Liu et al.}
\shorttitle{TA-SparseMG: Trend-Aware Sparse Forecasting via Multi-Scale Gating for Long-Term Time Series}

\title[mode = title]{TA-SparseMG: Trend-Aware Sparse Forecasting via Multi-Scale Gating for Long-Term Time Series} 

\author[1]{Wenchao Liu}
\credit{Conceptualization, Methodology, Software, Writing – original draft}

\author[2]{Hongbing Wang}
\credit{Supervision, Writing – review \& editing}

\author[1]{Youji Zhu}
\credit{Visualization, Software}

\author[2]{Xiaodong Liu}
\credit{Visualization, Software}

\author[2]{Xiangguang Xiong\corref{cor1}}
\credit{Conceptualization, Methodology, Supervision, Writing – review \& editing}
\ead{xxg@gznu.edu.cn}

\address[1]{School of Mathematical Sciences, Guizhou Normal University, Guiyang 550025, China.} 
\address[2]{School of Big Data and Computer Science, Guizhou Normal University, Guiyang 550025, China.}

\cortext[cor1]{Corresponding author}  

\begin{abstract}
Long-term time series forecasting finds extensive applications in domains such as power demand, traffic flow, meteorological observation, and renewable energy dispatch. Forecasting dynamically varying long-term time series poses inherent challenges, including statistical nonstationarity, local high-frequency disturbances, and coupled cross-period dependencies, which make it difficult for lightweight models to balance parameter efficiency and forecasting performance. To address this issue, this study presents TA-SparseMG, a lightweight cross-period forecasting model built on SparseTSF's sparse cross-period modeling framework. It incorporates three key modules: a trend-aware reversible instance normalization module, a scale-adaptive gated denoising module, and a multiscale gated-attention MLP forecasting module. The trend-aware normalization module captures input-window statistics and calibrates forecast-window distributions, effectively mitigating distribution shift. The scale-adaptive gated denoising module performs feature smoothing and residual suppression before period rearrangement, thereby reducing interference from high-frequency perturbations. The multiscale gated attention prediction module strengthens the prediction head's adaptive representational capacity via conditional gating and feature modulation. Extensive experiments across multiple LTSF benchmarks demonstrate that the proposed TA-SparseMG consistently achieves superior, stable performance. Ablation studies confirm that each module independently improves distribution adaptation, input robustness, and cross-period feature mapping capability.
\end{abstract}


\begingroup
\hfuzz=130pt
\maketitle
\endgroup

\newenvironment{fequation}{\footnotesize\begin{equation}}{\end{equation}}

\section{Introduction}
Long-term time series forecasting (LTSF) infers future long-horizon trends from historical observations. It is widely used in applications such as electricity load estimation, traffic flow analysis, weather forecasting, and renewable energy dispatch. Compared with short-term forecasting, LTSF requires a wider receptive field and the ability to cope with intricate periodic structures and pronounced shifts in statistical distributions, thereby imposing stringent requirements on model representational capacity, robustness, and computational efficiency. With advances in multivariate sensing and online monitoring, real-world time series have grown progressively longer and exhibit stronger nonstationarity, multiscale patterns, and cross-period properties. Effectively exploiting contextual information in long-term time series (LTS) while maintaining a compact parameter budget and balancing forecasting accuracy with stability remains a core research challenge in LTSF \cite{ref1}.

Existing LTSF approaches can be broadly classified into three technical lines: lightweight methods centered on linear transformations and frequency-domain fitting, Transformer-based methods for modeling long-range dependencies, and cross-period modeling methods that incorporate periodicity priors \cite{ref2}. Models such as DLinear \cite{ref3} and FITS \cite{ref4} adopt trend-seasonal decomposition and frequency-domain fitting to achieve competitive forecasting performance with minimal parameters. PatchTST \cite{ref5}, iTransformer \cite{ref6}, FEDformer \cite{ref7}, FreEformer \cite{ref8}, and TQNet \cite{ref9} focus on optimizing long-range dependency capture, inter-variable correlation mining, and frequency-domain feature extraction. SparseTSF \cite{ref10} and SimpleTM \cite{ref11} embed periodic priors to build lightweight cross-period models, achieving a favorable balance between computational cost and forecasting performance. Additionally, techniques such as distribution shift suppression \cite{ref12,ref13,ref14}, multiscale context modeling \cite{ref15,ref16,ref17,ref18}, and gated feature optimization \cite{ref19,ref20,ref21} offer viable pathways to further improve overall model performance. Collectively, existing studies have advanced LTSF across dimensions such as parameter optimization, dependency modeling, frequency-domain feature extraction, and the exploitation of periodicity.

Lightweight cross-period models have garnered growing attention due to their favorable efficiency-performance trade-off. By reconstructing time-series structures with periodic priors and conducting predictive modeling in a period-aligned feature space, these methods avoid computationally expensive global operations on the full time series. SparseTSF \cite{ref10}, for instance, discards elaborate attention mechanisms and streamlines modeling via periodic rearrangement, achieving state-of-the-art performance among lightweight models with limited parameters. Nevertheless, its performance still has room for improvement in handling dynamic LTSF tasks. In summary, existing methods confront three primary limitations. First, most normalization modules rely on static statistics from historical time series to align distributions, thereby failing to capture intra-series statistical dynamics effectively. When mean and volatility shift across historical and forecast windows, the model suffers from distribution mismatch, which degrades prediction reliability \cite{ref12,ref13,ref14}. Second, front-end time series processing lacks dedicated denoising mechanisms. High-frequency perturbations in raw data propagate into the feature space through periodic rearrangement, disrupting the learning of periodic patterns and undermining model stability \cite{ref10,ref15,ref16,ref17,ref18}. Third, prediction heads built on conventional shallow networks and linear transformations follow fixed mapping rules. They cannot dynamically adapt their computational logic to cross-period time-series patterns and thus struggle with scenarios involving multiscale variations, local noise, and complex periodic superpositions, thereby constraining their representational capacity \cite{ref3,ref10,ref19,ref20,ref21}. To address the above issues, this paper proposes TA-SparseMG, a lightweight enhanced cross-period model. Its main contributions are summarized as follows:

\begin{itemize}
\item Propose TA-SparseMG, a lightweight enhanced cross-period modeling framework. Built on SparseTSF's core sparse cross-period architecture, it retains the inherent parameter efficiency while introducing improvements in three dimensions: distribution adaptation, feature denoising, and prediction head optimization. The framework provides a balanced solution of forecasting accuracy, robustness, and computational efficiency for dynamic LTS tasks.
\item Design a trend-aware reversible instance normalization (TA-RevIN) module and a scale-adaptive gated denoising module. The former captures intra-series mean and volatility drift patterns and integrates them with estimated statistics to mitigate distribution discrepancy across windows. The latter performs time-series smoothing and residual suppression before periodic rearrangement, thereby alleviating the interference of high-frequency noise on periodic features.
\item Develop a multiscale gated attention MLP predictor. It optimizes feature transformation via conditional gating and attention mechanisms, enabling adaptive adjustment of mappings to diverse periodic patterns.
\item Extensive experiments across six mainstream LTSF benchmarks demonstrate that TA-SparseMG consistently achieves performance gains. Ablation studies further verify the effectiveness of each module.
\end{itemize}

The rest of this paper is structured as follows. Section~\ref{sec:related_work} reviews recent advances in LTSF, covering lightweight models, cross-period modeling, normalization-denoising techniques, and adaptive prediction heads, and summarizes the limitations of existing approaches. Section~\ref{sec:method} elaborates on the overall architecture of TA-SparseMG and the implementation details of its core modules. Section~\ref{sec:experiments} presents the experimental setup, benchmark results, and ablation analysis to validate the effectiveness of the proposed model and its components. Section~\ref{sec:conclusion} concludes the work, discusses remaining limitations, and outlines future research directions.

\section{Related work}
\label{sec:related_work}
\subsection{Lightweight Modeling in LTSF}
LTSF, which predicts future time-series values from historical observations, is widely adopted across the energy, meteorology, transportation, and finance domains. Early studies predominantly rely on statistical models such as ARIMA \cite{ref22} and conventional machine learning algorithms. In contrast, the deep learning era has established a technical paradigm centered on recurrent neural networks, convolutional neural networks (CNNs), and Transformers \cite{ref22,ref23,ref24}. Convolutional architectures can construct effective long-term memory and represent a mainstream approach for temporal modeling \cite{ref25}. Despite their strong capacity to capture long-range dependencies and inter-variable correlations, complex models incur excessive parameter overhead and computational cost, rendering them unsuitable for resource-constrained and real-time forecasting scenarios. Accordingly, striking a balance between forecasting accuracy and model complexity has emerged as a core research objective in LTSF.

Transformer-based architectures and their variants dominate mainstream LTSF research. Informer \cite{ref26} adopts sparse attention to reduce computational overhead. Autoformer \cite{ref27} integrates time series decomposition with autocorrelation mechanisms to extract periodic features. Crossformer \cite{ref28} employs segment embeddings and a two-stage attention mechanism to jointly model temporal and variable correlations. FEDformer \cite{ref7} leverages frequency-domain transformations to improve computational efficiency. Pyraformer \cite{ref29} uses a multi-resolution pyramid structure to simplify the modeling of long-range dependencies. Such models typically contain millions of parameters, imposing substantial constraints on practical deployment. Accordingly, research attention has increasingly shifted toward lightweight architectural designs. DLinear \cite{ref3}, FITS \cite{ref4}, N-BEATS \cite{ref30}, and SparseTSF \cite{ref10} are built upon shallow MLPs, with high parameter efficiency and training stability as core design priorities. TiDE \cite{ref31} constructs a fully connected encoder-decoder architecture, verifying the feasibility of modeling nonlinear dependencies via simple dense networks. TSMixer \cite{ref19} exploits cross-dimensional information fusion to bypass elaborate attention operations, strengthening the learning of temporal features and variable associations. ModernTCN \cite{ref32} fully leverages the computational advantages of convolutional structures to enhance the model's representation of long- and short-term temporal patterns. These methods achieve LTSF with low parameter budgets via linear-transform-based frequency-domain fitting and lightweight nonlinear modules, delivering competitive performance on mainstream benchmarks.

Lightweight models feature simple architectures, stable training, and strong scalability, bearing considerable practical value for engineering deployment. Nevertheless, their minimalist design inherently constrains representational capacity. When processing dynamic data with nonstationary distributional shifts, local high-frequency noise, and multi-periodic characteristics, static linear transformations and shallow MLPs struggle to fully uncover underlying patterns. Retaining the merits of lightweight models while improving distribution adaptability, robustness to noise, and the representational performance of the prediction head is critical to advancing lightweight LTSF techniques.

\subsection{Cyclical Priors and Lightweight Cross-Period Modeling}
LTS generally exhibit strong periodicity, and exploiting explicit periodic priors is an effective strategy to improve long-horizon forecasting performance and modeling efficiency \cite{ref33}. Existing studies reformulate long-term forecasting as cross-period trend modeling by restructuring raw LTS into period-aligned cross-period representations. Representative methods in this paradigm include SparseTSF/MLP \cite{ref10} and SimpleTM \cite{ref11}: the former enables cross-period forecasting with minimal parameter overhead, and the latter augments feature modeling capacity via a lightweight backbone \cite{ref20}. By leveraging periodic priors, these approaches alleviate the challenges of modeling long-range dependencies and enable efficient extraction of contextual information from LTS with a compact parameter budget.

Existing lightweight cross-period modeling methods primarily focus on period-aligned reshaping strategies and backbone architecture design, yet devote insufficient attention to representation quality before and after rearrangement, distribution-adaptation properties, and the adaptiveness of the prediction mapping. When applied to complex LTS featuring nonstationarity, local perturbations, and multiscale fluctuations, a fixed single-period prior paired with a lightweight backbone often fails to guarantee robust forecasting performance. Recent studies on periodic drift correction, such as Correctformer \cite{ref34}, also indicate that explicitly modeling periodic drift is important for improving forecast robustness to evolving temporal patterns. Accordingly, preserving the efficiency merits of cross-period modeling while improving the model's adaptability to complex input patterns and cross-period variations remains a pressing research priority in this field.

\subsection{Normalization, Denoising, and Adaptive Prediction Heads}
LTSF performance is governed not only by core modeling capacity but also by input distribution drift, local noise, and the representational quality of the prediction head. In terms of normalization, reversible instance normalization techniques mitigate sample-level distribution discrepancies and improve adaptability to nonstationary time series \cite{ref12}, \cite{ref35}. Nevertheless, most existing methods restore distributions using static statistics derived from the input window, thereby neglecting intra-window statistical drift. In long-horizon forecasting scenarios, the statistical properties of future time series deviate substantially from those of the input window; static means and standard deviations fail to accommodate dynamic distribution shifts \cite{ref13}, \cite{ref36}.

For input feature refinement, LTS commonly contain high-frequency disturbances, short-term noise, and anomalous fluctuations. Unprocessed interference contaminates deep-level features and degrades model stability. Studies on frequency-domain MLPs indicate that frequency-domain feature learning effectively captures global dependencies and energy distribution characteristics \cite{ref15}. Time-frequency fusion models, such as DTFNet \cite{ref37}, further show that jointly exploiting temporal and spectral cues is beneficial for modeling non-stationary time series. Existing works have proposed optimization strategies, including time-domain smoothing, frequency-domain filtering, time-series decomposition, and multiscale context modeling \cite{ref16,ref17,ref18,ref32,ref38}. Yet, these have not been systematically integrated into lightweight cross-period backbone architectures.

Regarding prediction head design, conventional shallow MLPs and linear mappings achieve high parameter efficiency but apply identical transformations to all samples and lack adaptive modulation for diverse temporal patterns \cite{ref19}, \cite{ref20}. Gating mechanisms and attention modulation can effectively enhance the pattern adaptability of prediction heads, but how to embed them into lightweight cross-period modeling frameworks at low computational cost remains under-explored.

Motivated by the above observations, this paper proposes TA-SparseMG, a lightweight cross-period backbone model optimized across three dimensions: distribution adaptation, input representation purification, and prediction head enhancement. The TA-RevIN module explicitly models intra-window statistical drift; the scale-adaptive gated denoising module suppresses high-frequency disturbances before periodic rearrangement; and the multiscale gated attention predictor augments the shallow prediction head's capacity to model complex cross-period patterns. This architecture requires no modification to the original lightweight backbone. Instead, it addresses performance bottlenecks through front-end and back-end auxiliary modules, thereby overcoming existing limitations while preserving computational efficiency.

\section{Proposed Method}
\label{sec:method}
\subsection{Problem Definition}
The LTSF task focuses on extrapolating future time steps from historical time-series observations. Given a $C$-variate time series with a look-back window length $L$ and a prediction horizon $T$, the historical input time series is denoted as $X=\{x_{t-L+1},x_{t-L+2},\ldots,x_t\}\in\mathbb{R}^{L\times C}$. The LTSF task is formalized as learning a mapping function $f:\mathbb{R}^{L\times C}\rightarrow\mathbb{R}^{T\times C}$ such that the generated predicted time series aligns as closely as possible with the ground-truth future time series $Y\in\mathbb{R}^{T\times C}$.

Real-world LTS commonly exhibit periodic patterns, statistical nonstationarity, and multiscale fluctuations. To leverage inherent periodic priors, let $P$ denote the dominant period length. The look-back window and forecast horizon are partitioned into $N=\lfloor L/P\rfloor$ complete historical periods and $M=\lceil T/P\rceil$ future periods, respectively. Reconstructing the original time series via periodic alignment yields the cross-period representation $Z\in\mathbb{R}^{N\times P\times C}$, which transforms the original LTSF task into cross-period trend modeling across individual phase positions. Built on a lightweight architecture, the model accommodates critical temporal properties including distribution shifts, high-frequency disturbances, and cross-period dependencies, thus delivering accurate and robust LTSF performance.

\subsection{Overall Framework}
\begin{figure*}[t]
	\centering
	\includegraphics[width=0.98\textwidth]{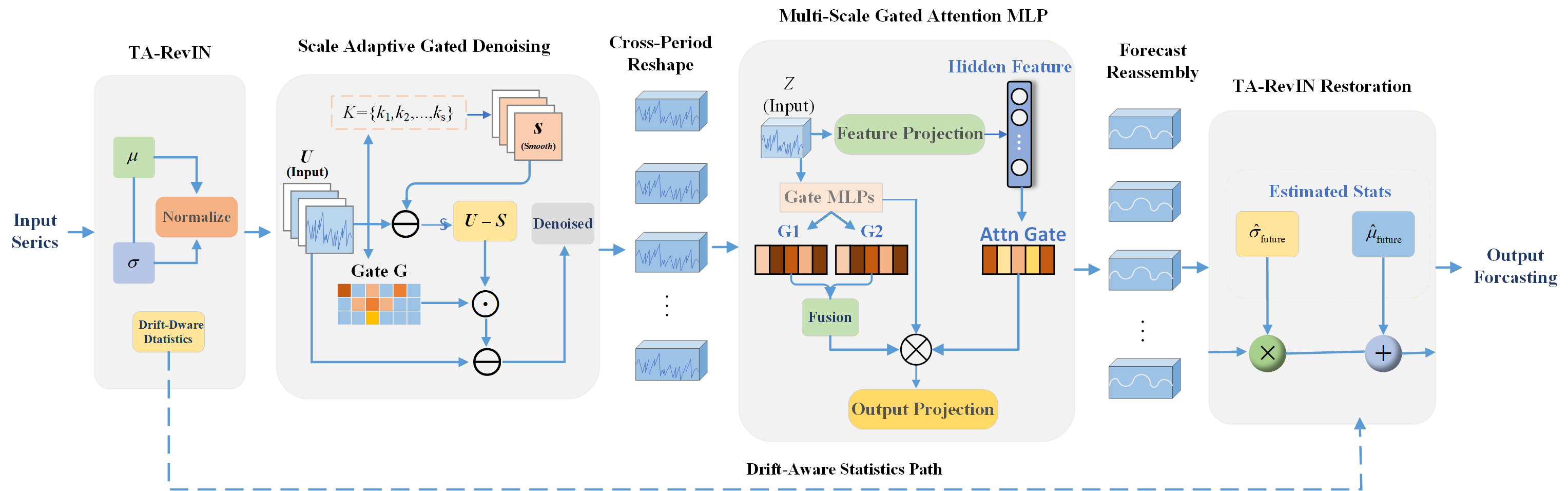}
	\caption{Overall framework of the proposed TA-SparseMG.}
	\label{fig:framework}
\end{figure*}

As illustrated in Fig.~\ref{fig:framework}, TA-SparseMG inherits the cross-period sparse modeling backbone from SparseTSF and establishes a complete inference pipeline comprising input preprocessing, periodic rearrangement, and prediction reconstruction. The TA-RevIN module first performs trend-aware normalization on the input time series to stabilize its temporal distribution. The scale-adaptive gated denoising module then extracts smooth trend components and suppresses high-frequency residual noise, yielding a purified time-series representation.

Guided by the dominant period length $P$, the model reshapes the denoised representations into period-aligned cross-period features and maps them to future period units via a multiscale gated-attention MLP predictor. Finally, the model rearranges the predicted outputs to recover the standard temporal structure and projects them back to the original data scale through the TA-RevIN denormalization module. Without altering the lightweight cross-period backbone, this architecture jointly improves the model's distribution adaptation capability, feature robustness, and cross-period forecasting performance.

\subsection{TA-RevIN module}
The left panel of Fig.~\ref{fig:framework} shows the proposed TA-RevIN module, which addresses statistical distribution shifts between historical and forecast windows in long-term time-series forecasting. Unlike conventional methods that perform normalization and restoration using static statistics derived solely from the input window, TA-RevIN integrates a trend-aware statistical branch into the sample-level reversible instance normalization framework. By modeling the dynamic drift of the mean and volatility within the input window, the module estimates statistical restoration parameters for the prediction window, thereby enhancing denormalization's adaptability to nonstationary time series. Specifically, for each input sample $X\in\mathbb{R}^{B\times L\times C}$, the module computes the per-variable mean $\mu$ and standard deviation $\sigma$ along the temporal dimension to conduct sample-wise normalization. It generates the normalized representation $X^{(n)}$ by incorporating learnable affine parameters $w$ and $b$.
\begin{equation}
X^{(n)}=\frac{X-\mu}{\sigma}\odot w+b,
\label{eq:ta_revin_norm}
\end{equation}
where $\odot$ denotes element-wise multiplication. This normalization scheme preserves the core framework of standard RevIN~\cite{ref12}, eliminating sample-level statistical variations while retaining feature representational capacity via learnable affine transformations. Nevertheless, the denormalization procedure of standard RevIN relies on static mean and standard deviation computed over the input window, which limits its adaptability to statistical distribution drift between historical and forecast windows in long-term forecasting scenarios. To mitigate this limitation, TA-RevIN dynamically estimates denormalization statistics for the forecast window by capturing temporal statistical drift in the input time series, thereby replacing the conventional recovery strategy that relies on fixed, static statistics.

To model the temporal statistical variations within the input window, the input time series of length $L$ is uniformly partitioned along the temporal dimension into two consecutive segments: the preceding segment $X_{\mathrm{pre}}$ and the succeeding segment $X_{\mathrm{post}}$. The mean and standard deviation of each segment are calculated, based on which the mean drift $\Delta\mu$ and volatility drift $\Delta\sigma$ are formally defined as follows.
\begin{equation}
\left\{
\begin{aligned}
\Delta\mu
&=
\operatorname{Mean}_{t}\left(X_{\mathrm{post}}\right)
-
\operatorname{Mean}_{t}\left(X_{\mathrm{pre}}\right),\\
\Delta\sigma
&=
\operatorname{Std}_{t}\left(X_{\mathrm{post}}\right)
-
\operatorname{Std}_{t}\left(X_{\mathrm{pre}}\right),
\end{aligned}
\right.
\label{eq:drift_statistics}
\end{equation}
where $\operatorname{Mean}_{t}(\cdot)$ and $\operatorname{Std}_{t}(\cdot)$ denote the mean and standard deviation of the corresponding time-series segments along the temporal dimension, respectively. $\Delta\mu$ denotes the temporal shift in the time series mean within the input window, while $\Delta\sigma$ reflects the phasic variation in the time series' amplitude. Compared with global static statistics, this mechanism provides finer-grained statistical information on the evolution of the distribution recovery process.

Let $Y^{(o)}$ denote the prediction output in the normalized space. 
As this output retains the affine transformation component introduced during normalization, 
The denormalization procedure first undoes the effect of the affine parameters, yielding the intermediate prediction feature $Y_{\epsilon}$.
\begin{equation}
Y_{\epsilon}
=
\frac{Y^{(o)}-b}{w+\epsilon},
\label{eq:remove_affine}
\end{equation}
where $\epsilon$ denotes a small positive constant introduced to prevent division by zero errors.

TA-RevIN then performs denormalization using the estimated recovery statistics of the forecast window. Specifically, $\hat{\mu}_{\mathrm{future}}$ denotes the mean recovery coefficient, and $\hat{\sigma}_{\mathrm{future}}$ denotes the standard deviation recovery coefficient. The final prediction result $\hat{Y}$ is obtained as
\begin{equation}
\hat{Y}
=
Y_{\epsilon}\odot \hat{\sigma}_{\mathrm{future}}
+
\hat{\mu}_{\mathrm{future}}.
\label{eq:denormalization}
\end{equation}

Note that $\hat{\mu}_{\mathrm{future}}$ and $\hat{\sigma}_{\mathrm{future}}$ are not the ground-truth mean and standard deviation of the time series, but rather recovery parameters derived from input-window statistics and internal drift estimation~\cite{ref12}. The mean recovery term enables compensation only when the drift magnitude exceeds a predefined threshold, preventing mild random disturbances from being misclassified as genuine trend shifts. The standard deviation recovery term integrates current volatility and drift characteristics to estimate the volatility range over the forecast window. The detailed calculations are presented as follows.

\begin{equation}
\left\{
\begin{aligned}
\hat{\mu}_\mathrm{future}
&= \mu + m_u \odot \Delta\mu \odot
\mathbb{I}\left(|\Delta\mu|>\tau_u\right), \\
\hat{\sigma}_\mathrm{future}
&= \max\left(\sigma + m_\sigma \odot \Delta\sigma,\epsilon\right),
\end{aligned}
\right.
\label{eq:future_statistics}
\end{equation}
where $m_u$ and $m_\sigma$ denote learnable drift coefficients, $\tau_u$ is the mean drift threshold, and $\mathbb{I}(\cdot)$ denotes the element-wise indicator function. When significant mean drift is detected, the mechanism adaptively adjusts the mean-recovery term for the forecast window. Combined with dynamic modulation of scale parameters via volatility drift, this design effectively enhances the denormalization procedure's adaptability to nonstationary time series.

As illustrated in Fig.~\ref{fig:framework}, the left normalization branch of TA-RevIN produces stable input representations. In contrast, the right denormalization branch reconstructs final predictions using the estimated parameters $\hat{\mu}_\mathrm{future}$ and $\hat{\sigma}_\mathrm{future}$. The bottom drift-aware statistical pathway exclusively conveys mean and volatility drift information; it does not participate in the main feature flow and is dedicated to computing statistical recovery parameters for the forecast window. By explicitly modeling intra-window statistical drift and optimizing denormalization through adaptive estimation of future statistics, the module establishes a reliable mapping for nonstationary LTS while preserving distributional stability across forecasting stages.

\subsection{Scale-Adaptive Gated Denoising}
SparseTSF employs lightweight one-dimensional convolutions for local information aggregation, yet such operations are restricted to intra-period feature modeling. This limitation impairs the model's ability to distinguish underlying temporal trends from high-frequency disturbances, and the framework lacks an adaptive mechanism to suppress residual noise. In LTSF tasks, short-term anomalies and high-frequency noise propagate into the cross-period representation space during periodic rearrangement, hampering the learning of inter-period dependency patterns. To address this issue, this work embeds a scale-adaptive gated denoising module upstream of the cross-period rearrangement pipeline. By integrating multiscale smoothing, residual decomposition, and gated control mechanisms, the module effectively enhances the stability and robustness of temporal representations.

Let $U$ denote the normalized channel-wise representation. This work constructs a set of multiscale smoothing convolutional kernels $K=\{k_1,k_2,\ldots,k_s\}$, where $k_i$ denotes the kernel size of the $i$-th smoothing branch, and $s$ is the number of smoothing branches. Each branch extracts channel-wise, multiscale, locally smoothed features along the temporal dimension, and the multi-branch outputs are fused via learnable weights to yield the global trend representation $S$. In parallel, a dedicated gating branch generates the gating map $G$, as formulated below.
\begin{equation}
\left\{
\begin{aligned}
S &= \sum_{i=1}^{s} a_i\,\operatorname{Conv}_{k_i}(U),\\
G &= \sigma\left(\operatorname{DWConv}(U)\right),
\end{aligned}
\right.
\label{eq:scale_adaptive_gated_denoising}
\end{equation}
where $a_i$ denotes the fusion weight of the $i$-th branch, $\operatorname{Conv}_{k_i}(\cdot)$ represents a one-dimensional smoothing convolution with kernel size $k_i$, $\operatorname{DWConv}(\cdot)$ is a channel-wise one-dimensional convolution that produces channel-level gating responses, and $\sigma(\cdot)$ denotes the Sigmoid activation function. Once the global trend representation $S$ is obtained, the local high-frequency residual of the time series is defined as the difference between the input representation and the smoothed trend.
\begin{equation}
R = U - S.
\label{eq:residual_definition}
\end{equation}

The residual term $R$ captures deviations from the smoothed trend, encoding high-frequency time-series variations that include both informative local details and undesirable perturbations, such as high-frequency noise and transient anomalies. Rather than simply removing the residual component, the model applies the gating map $G$ to selectively suppress it. The final denoised representation is formulated as
\begin{equation}
U_d = U - \lambda\left(R\odot G\right),
\label{eq:denoised_representation}
\end{equation}
where $\lambda$ denotes the constrained residual fusion coefficient that controls the intensity of denoising correction. Instead of directly replacing the original input $U$ with the smoothed trend $S$, the mechanism applies controlled adjustments to high-frequency residuals while retaining the core temporal structure. This design avoids feature detail loss induced by over-smoothing and mitigates the adverse impact of short-term disturbances on cross-period modeling.

The multi-kernel smoothing branch extracts multiscale local trends, adapting to temporal patterns of variation across diverse datasets and prediction horizons. The gated residual suppression strategy adaptively tunes the correction magnitude based on input characteristics, overcoming the limitations of fixed filtering schemes to achieve a dynamic trade-off between noise suppression and informative feature retention. The refined denoised representation $U_d$ is fed into the cross-period rearrangement module to generate period-aligned features. Compared with directly rearranging normalized temporal features on a per-period basis, this preprocessing feature-purification pipeline prevents high-frequency disturbances from propagating into the cross-period feature space, effectively enhancing the model's forecasting robustness.

\subsection{Multiscale Gated Attention MLP}
Following trend-aware normalization and gated denoising preprocessing, the model adopts the cross-period sparse backbone of SparseTSF. It reconstructs the input into a period-aligned cross-period representation based on the dominant period length $P$, then predicts future period units in this feature space. Let $U_d\in\mathbb{R}^{B\times C\times L}$ denote the denoised channel-first representation, $N=L/P$ denote the number of historical periods, and $M=T/P$ denote the number of future periods. The model first performs cross-period rearrangement to yield the global cross-period representation $Z$, which aggregates historical trajectories across all samples, variables, and period phases. A single cross-period unit $z$ extracted from this representation captures the temporal trajectory of consecutive historical periods at a fixed phase position. This operation transforms the LTSF task into a cross-period trend prediction task across different phase positions. By leveraging the prior periodic structure, it alleviates the difficulty of modeling long-range dependencies while retaining the lightweight, high-efficiency advantages of the SparseTSF framework.

The original SparseTSF/MLP~\cite{ref10} and MLP predictors map the historical period dimension $N$ of the cross-period representation $Z$ to the future period dimension $M$ via shared linear layers or shallow MLPs. Despite high parameter efficiency, this architecture applies a uniform nonlinear mapping to all representation units and lacks adaptive modulation of cross-period patterns. Consequently, its representational capacity is constrained in scenarios with coexisting multiscale fluctuations, local perturbations, and complex cross-period dependencies. To address this limitation, this work introduces a multiscale gated attention MLP predictor to enhance the adaptive modeling capability of the cross-period prediction head. Let $z\in\mathbb{R}^{N}$ denote a cross-period input representation unit extracted from $Z$, and let $\hat{z}\in\mathbb{R}^{M}$ denote the corresponding future-period representation. A hidden representation is first derived via feature projection.
\begin{equation}
h=\operatorname{Dropout}\left(\operatorname{ReLU}\left(W_1z+b_1\right)\right),
\label{eq:feature_projection}
\end{equation}
where $W_1$ and $b_1$ denote the parameters of the input projection layer, $\operatorname{ReLU}(\cdot)$ represents a nonlinear activation function, and $\operatorname{Dropout}(\cdot)$ mitigates overfitting. This step corresponds to the basic nonlinear mapping of the vanilla MLP prediction head.

Building on this foundation, the proposed model introduces a dual-path gated MLP that generates input-conditioned gates from the input representation unit $z$.
\begin{equation}
\left\{
\begin{aligned}
g_1 &= \sigma\left(W_{g1}z+b_{g1}\right),\\
g_2 &= \sigma\left(W_{g2}z+b_{g2}\right),
\end{aligned}
\right.
\label{eq:dual_gate}
\end{equation}
where $\sigma(\cdot)$ denotes the Sigmoid function. The two gated branches are subsequently fused via a learnable fusion coefficient to yield a composite gated representation.
\begin{equation}
g=\rho g_1+(1-\rho)g_2,
\label{eq:gate_fusion}
\end{equation}
where $g_1$ and $g_2$ denote the outputs of the dual parallel gated branches, and $\rho$ is the learnable fusion weight. Notably, the proposed multiscale mechanism does not employ multi-cycle parallel reconstruction. Instead, it applies differentiated conditional modulation to the same cross-period representation via multi-path gating, enhancing the model's adaptability to diverse temporal variation patterns.

An attention modulation factor is further derived from the hidden representation $h$ through attention gating.
\begin{equation}
a=\sigma\left(W_ah+b_a\right),
\label{eq:attention_gate}
\end{equation}
where $a$ quantifies the importance weight of each response in the hidden-layer features. The input-conditioned gating factor $g$ and the hidden attention modulation factor $a$ jointly adaptively modulate the hidden representations.
\begin{equation}
h_s=h+m_s\cdot\left(g\odot a\right),
\label{eq:hidden_modulation}
\end{equation}
where $m_s$ denotes a learnable scaling parameter that controls the modulation strength of the gating and attention mechanisms applied to hidden features, and $\odot$ denotes element-wise multiplication. This mechanism retains core features while adaptively amplifying relevant information in response to input patterns and hidden states.

The modulated features are subsequently projected to the future-period representation via the output layer.
\begin{equation}
\hat{z}=W_2h_s+b_2.
\label{eq:output_projection}
\end{equation}
The cross-period future representation $\hat{Z}$ is obtained by concatenating and fusing outputs across all samples, variables, and period-phase dimensions. It is then mapped back to the temporal dimension via predictive rearrangement, followed by distribution recovery performed by the TA-RevIN module. The proposed method inherits SparseTSF's cross-period sparse modeling backbone and converts LTSF into cross-period trend fitting by exploiting period-phase alignment. The core enhancement lies in the architecture of the prediction head. The multiscale gated attention predictor incorporates input-conditioned gating and latent-feature modulation while preserving lightweight mapping properties, enabling adaptive adjustment of mapping strategies to account for pattern variations across cross-period representations. This design effectively strengthens the modeling capacity for complex cross-period temporal patterns while retaining SparseTSF's merits of parameter efficiency and a prior periodic structure.

\subsection{Loss Function}
Following standard LTSF experimental settings, the proposed model employs mean squared error (MSE) as its loss function. Let $\hat{Y}\in\mathbb{R}^{B\times T\times C}$ denote the model's forecast and $Y\in\mathbb{R}^{B\times T\times C}$ denote the ground-truth future time series. The loss function is defined as
\begin{equation}
\mathcal{L}_{\mathrm{MSE}}
=
\frac{1}{\Omega}
\sum_{i=1}^{\Omega}
\left(y_i-\hat{y}_i\right)^2,
\label{eq:mse_loss}
\end{equation}
where $y_i$ and $\hat{y}_i$ denote the $i$-th ground-truth value and predicted value in the flattened time series, respectively, and $\Omega=B\times T\times C$ denotes the total number of scalar elements across all samples, forecasting steps, and variable dimensions. This objective computes the mean squared error over all scalar entries, directly quantifying the overall deviation between model predictions and ground-truth time series.

\section{Experiment and Analysis}
\label{sec:experiments}
\subsection{Experimental Setup}
\textbf{Datasets:} To evaluate the LTSF performance of TA-SparseMG, this work selects six mainstream LTSF benchmarks for experiments: ETTh1, ETTh2, Weather, Electricity, Solar-Energy, and Traffic. Covering typical industrial scenarios spanning energy, transportation, meteorology, and renewable energy, these datasets feature diverse sampling frequencies, variable dimensions, and periodic properties, enabling a comprehensive assessment of the model's generalization across different tasks. ETTh1 and ETTh2 are publicly available datasets for power transformer temperature monitoring~\cite{ref26}. The Weather dataset contains rich seasonal meteorological features. The Electricity and Traffic datasets record dynamic variations in power load and traffic flow, respectively. The Solar-Energy dataset is adopted to assess the model's adaptability to renewable energy output forecasting.

\textbf{Parameter Settings:} Experimental configurations were standardized to ensure fair comparisons across methods. The input look-back window length is fixed at $L=720$. Four prediction horizons, i.e., $T\in\{96,192,336,720\}$, are set to comprehensively evaluate the model's adaptability and stability across different forecasting ranges. All compared methods follow identical training and testing protocols. For each dataset, hyperparameters were selected from a small candidate set according to the validation loss. The test set was used only for the final evaluation. For high-dimensional datasets such as Solar-Energy and Traffic, a slightly larger training budget was allowed due to their slower validation convergence.

\textbf{Baselines:} Nine representative LTSF methods were selected as comparison baselines: DLinear~\cite{ref3}, FITS~\cite{ref4}, PatchTST~\cite{ref5}, iTransformer~\cite{ref6}, FEDformer~\cite{ref7}, TQNet~\cite{ref9}, SparseTSF/MLP~\cite{ref10},  SimpleTM~\cite{ref11}, and FiLM~\cite{ref16}. The selected baselines cover linear lightweight models, frequency-domain modeling approaches, Transformer-based architectures, and cross-period lightweight modeling paradigms, enabling comprehensive verification of the proposed model's performance advantages across different technical lines.

\textbf{Experimental Environment:} Both the proposed TA-SparseMG and all baseline models were implemented in PyTorch. All experiments were run on a single NVIDIA RTX 4090 GPU equipped with 24 GB of VRAM.

\textbf{Evaluation Metrics and Loss Functions:} The MSE and mean absolute error (MAE) were adopted as core evaluation metrics to quantify the model's predictive performance.
\begin{equation}
\left\{
\begin{aligned}
\mathrm{MSE}
&=
\frac{1}{N_{\mathrm{test}}}
\sum_{i=1}^{N_{\mathrm{test}}}
\left(y_i-\hat{y}_i\right)^2,\\
\mathrm{MAE}
&=
\frac{1}{N_{\mathrm{test}}}
\sum_{i=1}^{N_{\mathrm{test}}}
\left|y_i-\hat{y}_i\right|,
\end{aligned}
\right.
\label{eq:evaluation_metrics}
\end{equation}
where $\hat{y}_i$ denotes the predicted value, $y_i$ denotes the ground-truth value, and $N_{\mathrm{test}}$ denotes the total number of test samples. Lower MSE and MAE values correspond to superior forecasting performance. This work trains and evaluates all models under identical data splits and prediction configurations. The proposed method uses MSE as the training loss and reports both MSE and MAE on the test set.

\subsection{Main Results}
Table~\ref{tab:main_results} summarizes the quantitative comparison of TA-SparseMG against multiple baseline models across six datasets and four forecasting horizons. Performance is evaluated using $\mathrm{MSE}$ and $\mathrm{MAE}$, where lower values indicate higher forecasting accuracy. The best and second-best results are highlighted in red and blue, respectively. Experimental results demonstrate that the proposed model achieves favorable performance across most datasets and forecasting horizons.

Compared with the baseline SparseTSF/MLP~\cite{ref10}, the proposed model achieves lower $\mathrm{MSE}$ in $22/24$ settings and lower $\mathrm{MAE}$ in $24/24$ test tasks. It yields consistent performance improvements across the vast majority of experimental settings, with particularly notable gains on the ETTh1, ETTh2, and Weather datasets.

Furthermore, the proposed model maintains stable forecasting performance on high-dimensional and complex datasets, including Electricity, Solar-Energy, and Traffic. High dimensionality, complex inter-variable correlations, and severe local fluctuations characterize these datasets. Despite these challenges, TA-SparseMG remains competitive with strong baselines across most forecasting horizons, demonstrating its excellent adaptability to diverse scenarios. The result indicates that TA-SparseMG not only effectively improves the predictive performance of the original SparseTSF/MLP~\cite{ref10} but also generalizes well across a wide range of LTSF tasks.

In summary, TA-SparseMG achieves performance improvements without increasing the number of parameters. While retaining the lightweight cross-period backbone, the model effectively strengthens its modeling capabilities for nonstationary temporal characteristics, local noise disturbances, and complex cross-period patterns through three functional modules: trend-aware normalization and restoration, scale-adaptive gated denoising, and a multiscale gated-attention prediction head.

\begin{table*}[t]
\centering
\refstepcounter{table}\label{tab:main_results}
\makebox[\textwidth][l]{%
\begin{minipage}{\textwidth}
\small\textbf{Table~\thetable}\par
Comparison of multivariate LTSF results between our model and other mainstream baselines. The best results are highlighted in red, and the second-best results are highlighted in blue with underlines. The look-back length $L$ is uniformly set to 720, and the forecast horizon $T \in \{96,192,336,720\}$. Model publication years are shown in parentheses.
\end{minipage}}
\par\smallskip
\setlength{\tabcolsep}{2.4pt}
\renewcommand{\arraystretch}{1.18}
\papertablestyle

\resizebox{\textwidth}{!}{%
\begin{tabular}{
c|c|
c@{\hspace{6pt}}c|
c@{\hspace{6pt}}c|
c@{\hspace{6pt}}c|
c@{\hspace{6pt}}c|
c@{\hspace{6pt}}c|
c@{\hspace{6pt}}c|
c@{\hspace{6pt}}c|
c@{\hspace{6pt}}c|
c@{\hspace{6pt}}c|
c@{\hspace{6pt}}c
}
\hline
\multicolumn{2}{c|}{\textbf{Models}}
& \multicolumn{2}{c|}{\textbf{Ours}}
& \multicolumn{2}{c|}{\longmodel{SparseTSF}{/MLP}{2026}\cite{ref10}}
& \multicolumn{2}{c|}{\model{SimpleTM}{2025}\cite{ref11}}
& \multicolumn{2}{c|}{\model{TQNet}{2025}\cite{ref9}}
& \multicolumn{2}{c|}{\model{FITS}{2024}\cite{ref4}}
& \multicolumn{2}{c|}{\model{iTransformer}{2024}\cite{ref6}}
& \multicolumn{2}{c|}{\model{PatchTST}{2023}\cite{ref5}}
& \multicolumn{2}{c|}{\model{DLinear}{2023}\cite{ref3}}
& \multicolumn{2}{c|}{\model{FEDformer}{2022}\cite{ref7}}
& \multicolumn{2}{c}{\model{FiLM}{2022}\cite{ref16}} \\[2pt]
\hline
\multicolumn{2}{c|}{\textbf{Metric}}
& \textbf{MSE} & \textbf{MAE}
& \textbf{MSE} & \textbf{MAE}
& \textbf{MSE} & \textbf{MAE}
& \textbf{MSE} & \textbf{MAE}
& \textbf{MSE} & \textbf{MAE}
& \textbf{MSE} & \textbf{MAE}
& \textbf{MSE} & \textbf{MAE}
& \textbf{MSE} & \textbf{MAE}
& \textbf{MSE} & \textbf{MAE}
& \textbf{MSE} & \textbf{MAE} \\
\hline

\dataset{ETTh1}
& 96  & \first{0.354} & \first{0.384} & \second{0.372} & \second{0.395} & 0.383 & 0.421 & 0.377 & 0.405 & 0.382 & 0.405 & 0.396 & 0.426 & 0.375 & 0.404 & 0.381 & 0.407 & 0.439 & 0.479 & 0.610 & 0.582 \\ \cline{2-22}
& 192 & \first{0.398} & \first{0.411} & \second{0.412} & \second{0.417} & 0.419 & 0.440 & 0.430 & 0.442 & 0.417 & 0.425 & 0.430 & 0.450 & 0.418 & 0.430 & 0.419 & 0.430 & 0.469 & 0.493 & 0.613 & 0.585 \\ \cline{2-22}
& 336 & \first{0.433} & \first{0.428} & 0.444 & \second{0.439} & 0.440 & 0.463 & 0.454 & 0.456 & \second{0.436} & 0.442 & 0.480 & 0.486 & 0.462 & 0.460 & 0.454 & 0.456 & 0.521 & 0.528 & 0.633 & 0.603 \\ \cline{2-22}
& 720 & \second{0.453} & \second{0.468} & 0.499 & 0.497 & 0.476 & 0.495 & 0.504 & 0.508 & \first{0.433} & \first{0.455} & 0.700 & 0.608 & 0.495 & 0.497 & 0.499 & 0.512 & 0.624 & 0.587 & 0.769 & 0.684 \\
\hline

\dataset{ETTh2}
& 96  & 0.294 & 0.355 & 0.304 & 0.359 & \second{0.285} & \second{0.348} & 0.295 & 0.358 & \first{0.272} & \first{0.336} & 0.311 & 0.363 & 0.290 & 0.354 & 0.302 & 0.367 & 0.398 & 0.449 & 0.446 & 0.464 \\ \cline{2-22}
& 192 & 0.347 & 0.390 & 0.359 & 0.397 & \first{0.331} & \second{0.381} & 0.377 & 0.404 & \second{0.333} & \first{0.375} & 0.392 & 0.414 & 0.382 & 0.419 & 0.411 & 0.437 & 0.412 & 0.457 & 0.457 & 0.469 \\ \cline{2-22}
& 336 & 0.367 & 0.413 & 0.387 & 0.424 & \first{0.345} & \second{0.397} & 0.379 & 0.414 & \second{0.355} & \first{0.396} & 0.415 & 0.438 & 0.421 & 0.448 & 0.542 & 0.514 & 0.440 & 0.474 & 0.480 & 0.499 \\ \cline{2-22}
& 720 & 0.404 & 0.442 & 0.423 & 0.460 & 0.405 & \second{0.440} & 0.427 & 0.455 & \first{0.378} & \first{0.423} & 0.425 & 0.455 & \second{0.404} & 0.444 & 0.888 & 0.670 & 0.469 & 0.506 & 0.528 & 0.527 \\
\hline

\dataset{Electricity}
& 96  & \second{0.132} & \first{0.225} & 0.135 & \second{0.227} & \second{0.132} & \second{0.227} & 0.135 & 0.231 & 0.145 & 0.248 & 0.133 & 0.229 & \first{0.130} & \first{0.225} & 0.136 & 0.236 & 0.255 & 0.362 & 0.216 & 0.330 \\ \cline{2-22}
& 192 & \second{0.147} & \first{0.239} & 0.149 & \second{0.240} & \first{0.142} & 0.242 & 0.150 & 0.245 & 0.159 & 0.260 & 0.152 & 0.249 & \second{0.147} & 0.241 & 0.150 & 0.249 & 0.288 & 0.387 & 0.213 & 0.330 \\ \cline{2-22}
& 336 & \second{0.162} & \first{0.255} & 0.165 & 0.259 & \first{0.161} & 0.259 & 0.167 & 0.263 & 0.175 & 0.275 & 0.169 & 0.266 & \second{0.162} & \second{0.257} & 0.166 & 0.266 & 0.310 & 0.405 & 0.215 & 0.333 \\ \cline{2-22}
& 720 & \second{0.199} & \first{0.287} & 0.202 & 0.291 & 0.208 & 0.305 & 0.201 & 0.296 & 0.212 & 0.305 & \first{0.194} & \second{0.288} & 0.201 & 0.293 & 0.200 & 0.299 & 0.323 & 0.410 & 0.232 & 0.346 \\
\hline

\dataset{Solar-Energy}
& 96  & 0.174 & \second{0.229} & \second{0.172} & 0.233 & 0.189 & 0.241 & 0.181 & 0.242 & 0.192 & 0.241 & 0.190 & 0.259 & \first{0.167} & \first{0.226} & 0.192 & 0.260 & 0.326 & 0.417 & 0.673 & 0.622 \\ \cline{2-22}
& 192 & 0.193 & \first{0.243} & \second{0.192} & 0.248 & 0.195 & 0.260 & 0.203 & 0.261 & 0.214 & 0.253 & 0.204 & 0.270 & \first{0.186} & \second{0.245} & 0.212 & 0.275 & 0.336 & 0.412 & 0.670 & 0.645 \\ \cline{2-22}
& 336 & \first{0.197} & \first{0.245} & 0.202 & \second{0.256} & \second{0.199} & 0.261 & 0.219 & 0.272 & 0.231 & 0.261 & 0.218 & 0.282 & 0.213 & 0.257 & 0.228 & 0.287 & 0.320 & 0.399 & 0.710 & 0.669 \\ \cline{2-22}
& 720 & \second{0.204} & \first{0.254} & 0.205 & 0.260 & \first{0.198} & \second{0.258} & 0.231 & 0.281 & 0.237 & 0.265 & 0.218 & 0.282 & 0.228 & 0.266 & 0.237 & 0.295 & 0.437 & 0.447 & 0.729 & 0.681 \\
\hline

\dataset{Traffic}
& 96  & \second{0.362} & \first{0.243} & 0.364 & \second{0.245} & \first{0.359} & 0.256 & 0.382 & 0.269 & 0.398 & 0.286 & \second{0.362} & 0.264 & 0.370 & 0.257 & 0.396 & 0.288 & 0.647 & 0.415 & 0.578 & 0.354 \\ \cline{2-22}
& 192 & \second{0.372} & \first{0.249} & 0.380 & \second{0.256} & \first{0.364} & 0.264 & 0.393 & 0.273 & 0.409 & 0.289 & 0.375 & 0.270 & 0.382 & 0.262 & 0.406 & 0.290 & 0.684 & 0.427 & 0.576 & 0.351 \\ \cline{2-22}
& 336 & \second{0.385} & \first{0.258} & 0.401 & 0.273 & \first{0.380} & 0.272 & 0.434 & 0.319 & 0.421 & 0.294 & 0.390 & 0.278 & 0.395 & \second{0.269} & 0.419 & 0.296 & 0.741 & 0.453 & 0.580 & 0.350 \\ \cline{2-22}
& 720 & \second{0.430} & \second{0.283} & 0.439 & \second{0.283} & \second{0.430} & \first{0.263} & 0.447 & 0.302 & 0.457 & 0.311 & \first{0.423} & 0.296 & 0.434 & 0.289 & 0.459 & 0.319 & 0.756 & 0.467 & 0.629 & 0.383 \\
\hline

\dataset{Weather}
& 96  & \first{0.146} & \first{0.199} & 0.155 & 0.210 & 0.151 & 0.204 & 0.155 & 0.209 & 0.170 & 0.225 & 0.180 & 0.232 & \second{0.149} & \second{0.201} & 0.169 & 0.228 & 0.347 & 0.410 & 0.501 & 0.513 \\ \cline{2-22}
& 192 & \first{0.191} & \first{0.242} & 0.199 & 0.254 & 0.194 & 0.246 & 0.212 & 0.259 & 0.212 & 0.260 & 0.226 & 0.267 & \first{0.191} & \second{0.244} & 0.214 & 0.272 & 0.351 & 0.402 & 0.512 & 0.546 \\ \cline{2-22}
& 336 & \first{0.239} & \first{0.270} & 0.248 & 0.291 & \second{0.243} & \second{0.284} & 0.260 & 0.299 & 0.258 & 0.294 & 0.286 & 0.312 & 0.245 & 0.288 & 0.256 & 0.306 & 0.415 & 0.446 & 0.560 & 0.567 \\ \cline{2-22}
& 720 & \second{0.310} & \second{0.332} & 0.323 & 0.347 & \first{0.309} & \first{0.331} & 0.318 & 0.336 & 0.320 & 0.339 & 0.353 & 0.357 & 0.315 & 0.336 & 0.315 & 0.354 & 0.409 & 0.432 & 0.589 & 0.582 \\
\hline

\end{tabular}%
}
\end{table*}

\subsection{Ablation Study}
To objectively quantify the effectiveness of each core module, this work performs ablation studies across the ETTh1, ETTh2, Traffic, and Weather datasets. Four model variants are designed for comparative analysis:

\textbf{Full Model:} The complete TA-SparseMG architecture with all proposed modules enabled.

\textbf{w/o TA-RevIN:} The TA-RevIN module is removed, and the original RevIN is adopted for data normalization and distribution recovery.

\textbf{w/o Denoising:} The scale-adaptive gated denoising module is omitted, and the normalized features are fed directly into the cross-period prediction backbone.

\textbf{w/o MSGA:} The multiscale gated-attention prediction head is replaced with the native shallow MLP predictor of SparseTSF/MLP~\cite{ref10}.

\begin{table}[t]
\centering
\caption{Ablation study on ETTh1, ETTh2, Traffic, and Weather. Full denotes the complete model. Each ablation removes one component. Lower $\mathrm{MSE}$ and $\mathrm{MAE}$ indicate better performance. The best results are highlighted in bold.}
\label{tab:ablation_study}
\papertablestyle
\setlength{\tabcolsep}{6pt}
\renewcommand{\arraystretch}{1.12}

\begin{tabular}{llll}
\hline
\textbf{Dataset} & \textbf{Ablation Variant} & \textbf{Avg. MSE} & \textbf{Avg. MAE} \\
\hline
\multirow{4}{*}{ETTh2}
& \textbf{Full}            & \textbf{0.353} & \textbf{0.400} \\
& w/o TA-RevIN        & 0.358              & 0.403          \\
& w/o MSGA             & 0.358              & 0.403          \\
& w/o Denoising       & 0.355              & 0.401          \\
\hline
\multirow{4}{*}{ETTh1}
& \textbf{Full}            & \textbf{0.409} & \textbf{0.423} \\
& w/o TA-RevIN        & 0.413              & \textbf{0.423} \\
& w/o MSGA             & 0.428              & 0.431          \\
& w/o Denoising       & 0.415             & 0.426          \\
\hline
\multirow{4}{*}{Traffic}
& \textbf{Full}            & \textbf{0.387} & \textbf{0.258} \\
& w/o TA-RevIN        & 0.390              & 0.261          \\
& w/o MSGA             & 0.396              & 0.264          \\
& w/o Denoising       & 0.390             & \textbf{0.258} \\
\hline
\multirow{4}{*}{Weather}
& \textbf{Full}            & \textbf{0.221} & \textbf{0.263} \\
& w/o TA-RevIN        & 0.222              & \textbf{0.263} \\
& w/o MSGA             & 0.224             & 0.266          \\
& w/o Denoising       & 0.224            & 0.264          \\
\hline
\end{tabular}
\end{table}

As shown in Table~\ref{tab:ablation_study}, the full TA-SparseMG model achieves the best or second-best performance across all four test datasets, confirming the functional contribution of each core module and the non-redundant design of the overall architecture. Removing individual modules yields consistent degradation trends in the average evaluation metrics. The full model records an average $\mathrm{MSE}$ of $0.343$. Ablating the TA-RevIN module, the scale-adaptive gated denoising module, and the MSGA prediction head increases the average $\mathrm{MSE}$ to $0.346$, $0.346$, and $0.352$, respectively. These findings indicate that the three modules respectively enable distribution-adaptive calibration, input feature purification, and predictive representation enhancement, with mutually complementary functional effects.

The scale-adaptive gated denoising module delivers consistent performance gains. Removing this module results in degradation in $\mathrm{MSE}$ across ETTh1, ETTh2, Traffic, and Weather. Applying noise suppression to temporal features before periodic rearrangement effectively mitigates the adverse impact of high-frequency disturbances on cross-period feature modeling. Ablation results for the TA-RevIN module further confirm that the trend-aware statistical recovery strategy enhances the model's adaptability to nonstationary temporal distributions and effectively addresses the prevalent mean shifts and volatility variations in LTSF tasks.

In summary, the performance improvements of TA-SparseMG arise from the synergistic optimization of the three functional modules, rather than the isolated enhancement of any single component. These results validate the effectiveness of optimizing lightweight cross-period backbones via distribution calibration, feature purification, and prediction-head enhancement.
\FloatBarrier

\subsection{Analysis of the MSGA Mechanism}
This section verifies the effectiveness of the MSGA module in addressing the inherent limitations of the original SparseTSF/MLP~\cite{ref10} prediction head. The native shallow MLP predictor in SparseTSF/MLP~\cite{ref10} applies a fixed nonlinear mapping to the input features and lacks adaptive scale modulation. This structural limitation becomes increasingly prominent in complex forecasting scenarios where multiple frequency components coexist. The proposed MSGA module integrates multiscale representation learning with a gated fusion mechanism. Its multi-branch architecture captures discriminative periodic-scale features and dynamically calibrates branch weights conditioned on input states. This design endows the static shallow prediction pipeline with scale adaptability and effectively enhances the model's capacity to model complex cross-periodic patterns.

This work conducts comparative experiments, grouped by spectral complexity, to quantitatively validate the performance improvements of MSGA. The complexity of each test sample is measured using normalized spectral entropy, and all samples are partitioned into three complexity groups, namely Low, Medium, and High, based on spectral entropy tertiles. To isolate the impact of the prediction head, the experiment replaces only the prediction module. It compares MSGA with the original two-layer MLP, keeping all other modules and hyperparameters fixed.

This work quantifies the spectral complexity of each test sample's forecast window using normalized spectral entropy \cite{ref39}. Specifically, a real-valued fast Fourier transform (i.e., rFFT) is applied to the ground-truth future time series $y$ along the temporal dimension to extract non-redundant frequency components. The energy distribution of each component is computed as the square of the amplitude of the corresponding frequency coefficient.
\begin{equation}
\left\{
\begin{aligned}
\gamma_k &= \operatorname{rFFT}(y)_k,\\
P_k &= |\gamma_k|^2,
\end{aligned}
\right.
\label{eq:spectral_energy}
\end{equation}
where $\gamma_k$ denotes the $k$-th frequency component obtained via rFFT, and $P_k$ represents the corresponding spectral energy. Normalizing the spectral energy values yields a valid probability distribution across frequency components.
\begin{equation}
p_k=
\frac{P_k}
{\sum_{j=1}^{K}P_j}.
\label{eq:spectral_probability}
\end{equation}
Subsequently, the normalized spectral entropy is computed as
\begin{equation}
H_{\mathrm{norm}}
=
-\frac{1}{\log K}
\sum_{k=1}^{K}p_k\log(p_k),
\label{eq:normalized_spectral_entropy}
\end{equation}
where $K$ denotes the total number of frequency components. Lower spectral entropy indicates a concentrated spectral energy distribution, suggesting that a single periodic pattern or simple temporal variations dominate the sample. Higher spectral entropy, by contrast, reflects energy dispersed across multiple frequency components, which implies complex multiscale temporal characteristics. Note that spectral entropy is used exclusively for post hoc grouping analysis of test samples and does not participate in model training or hyperparameter selection.

For each dataset and forecasting horizon setting, the test samples are divided into three complexity groups according to the tertiles of their spectral entropy. Let $q_1$ and $q_2$ denote the $1/3$ and $2/3$ quantiles of spectral entropy under the corresponding setting, respectively. The sample grouping is defined as $\text{Low}: H_{\mathrm{norm}}\le q_1,\ \text{Medium}: q_1 < H_{\mathrm{norm}}\le q_2,\ \text{High}: H_{\mathrm{norm}}>q_2$.

This work compares the prediction errors of the full model and a variant in which the MSGA module is replaced with a native shallow MLP across all complexity groups. To ensure controlled experimental conditions, the structural parameters of the TA-RevIN and scale-adaptive gated denoising modules remain fixed, and only the prediction head architecture is modified. The relative performance gain is formally defined as
\begin{equation}
\mathrm{Gain}
=
\frac{
\mathrm{MSE}_{\mathrm{w/o\ MSGA}}
-
\mathrm{MSE}_{\mathrm{Full}}
}
{
\mathrm{MSE}_{\mathrm{w/o\ MSGA}}
}
\times 100\%.
\label{eq:msga_gain}
\end{equation}

\begin{figure*}[t]
\centering
\includegraphics[width=0.98\textwidth]{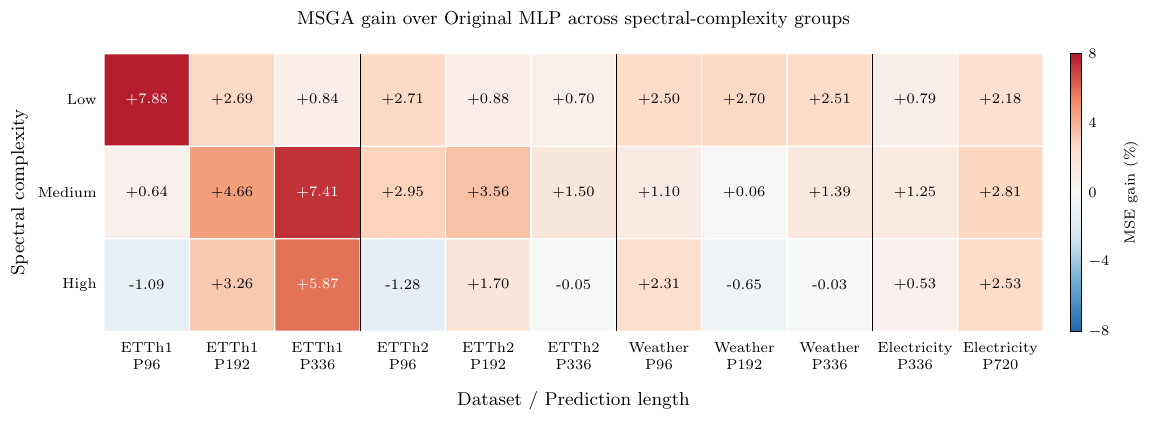}
\par\smallskip
\refstepcounter{figure}\label{fig:msga_gain}
\makebox[\textwidth][l]{%
\begin{minipage}{\textwidth}
\small\textbf{Fig.~\thefigure.}\enspace
Relative MSE gain of the full model over the original MLP predictor across spectral-complexity groups. Each column denotes a dataset-prediction-length setting, and each row denotes a spectral-complexity group. Positive values indicate that MSGA improves over the original MLP predictor, while negative values indicate degradation.
\end{minipage}}
\end{figure*}

Fig.~\ref{fig:msga_gain} shows that MSGA improves forecasting performance across most experimental configurations and generally outperforms the native two-layer MLP prediction head. The experiment covers 33 combinations of prediction horizons and spectral complexity groups, and the proposed module achieves an MSE reduction in 28 of these settings. For typical setups including ETTh1-P192, ETTh1-P336, ETTh2-P192, and Electricity-P720, the performance gains are more pronounced on samples with medium and high complexity. Although a few high-complexity groups show slight degradation, the differences are marginal. The result confirms that MSGA, through its scale-adaptive modulation mechanism, effectively addresses the static-mapping limitation of shallow MLPs, thereby enhancing the model’s capacity to capture complex temporal patterns.

\subsection{Analysis of Scale-Dependent Noise Cancellation Mechanisms}
To further analyze the mechanism of the scale-adaptive gated denoising module, this section quantifies its suppression effect on local high-frequency disturbances from a frequency-domain perspective \cite{ref7}, \cite{ref8}, \cite{ref31}. Short-term anomalies, high-frequency fluctuations, and random disturbances in LTS propagate into the cross-period representation space via periodic rearrangement, disrupting trend modeling in the periodic domain. Accordingly, reducing high-frequency energy in input representations before periodic rearrangement via the front-end denoising module yields more stable inputs for the subsequent cross-period prediction backbone.

Instead of comparing only pre- and post-processing features, this work identifies three critical intermediate representations within the scale-adaptive gated denoising module: the input representation $U$, the smoothed representation $S$ produced by the multiscale smoothing branch, and the final denoised representation $U_d$ after gated residual suppression. Specifically, $S$ reflects the multiscale smoothing branch's capability to extract low-frequency trends, and $U_d$ serves as the final output fed into the subsequent cross-period rearrangement module. Joint analysis of $U$, $S$, and $U_d$ thus reveals the module's internal working pipeline in detail: the multiscale smoothing branch first extracts trend-dominated low-frequency structures; the gating mechanism then selectively calibrates high-frequency residuals; the module finally outputs a denoised representation that strikes a balance between stationarity and local information retention.

For quantitative evaluation, this work computes rFFT along the temporal dimension for each variable channel of the test samples. It takes the squared amplitude of each frequency coefficient as the energy of the corresponding frequency component. Frequency components are sorted in ascending order, and the top $25\%$ are defined as the high-frequency band. The ratio of high-frequency energy to total spectral energy is then calculated for each representation.

\begin{equation}
\rho_{\mathrm{HF}}(X)
=
\frac{
\sum_{k\in\mathcal{H}}\left|F(X)_k\right|^2
}{
\sum_{k}\left|F(X)_k\right|^2
},
\label{eq:high_frequency_ratio}
\end{equation}
where $F(X)_k$ denotes the $k$-th frequency component of time series $X$ obtained via rFFT, and $\mathcal{H}$ denotes the high-frequency band consisting of the top $25\%$ of frequency components. To further quantify the high-frequency suppression effect at different processing stages, this work defines the relative reduction ratio of high-frequency energy from representation $A$ to representation $B$ as

\begin{equation}
\Delta_{\mathrm{HF}}(A\rightarrow B)
=
\frac{
\rho_{\mathrm{HF}}(A)-\rho_{\mathrm{HF}}(B)
}{
\rho_{\mathrm{HF}}(A)
}
\times 100\%,
\label{eq:high_frequency_reduction}
\end{equation}
where $A$ and $B$ denote the feature representations before and after the target processing stage, respectively. This work computes two core relative reduction ratios, $U\rightarrow S$ and $U\rightarrow U_d$. The former quantifies the low-pass filtering capacity of the multiscale smoothing branch. At the same time, the latter reflects the actual purification effect of the final denoised features, which serve as inputs to the subsequent cross-period backbone. All frequency-domain statistics are aggregated from nine test batches retained for each dataset--prediction length configuration.

\begin{figure*}[t]
\centering
\includegraphics[width=0.98\textwidth]{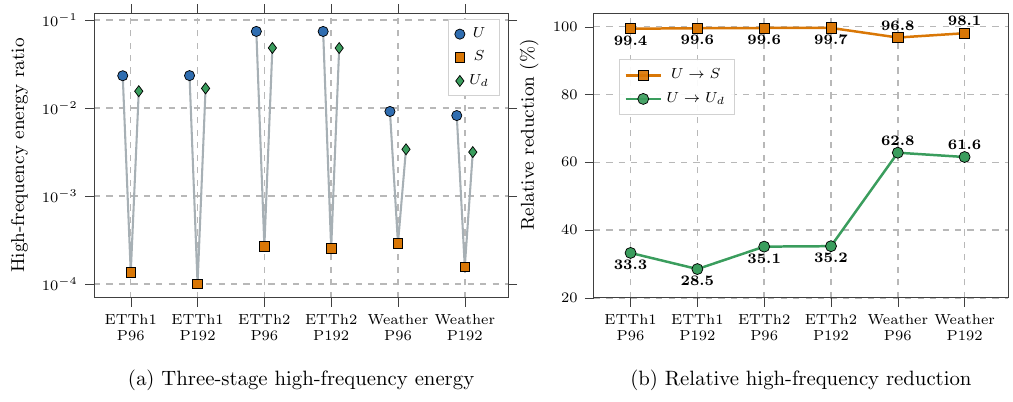}
\par\smallskip
\refstepcounter{figure}\label{fig:hf_energy}
\makebox[\textwidth][l]{%
\begin{minipage}{\textwidth}
\small\textbf{Fig.~\thefigure.}\enspace
High-frequency energy analysis of the scale-adaptive gated denoising module. (a) High-frequency energy ratios for the input representation $U$, the filtered representation $S$, and the final denoised output $U_d$. (b) Relative reduction of the ratio from $U$ to $S$ and from $U$ to $U_d$. The high-frequency region is defined as the top 25\% of the frequency components.
\end{minipage}}
\end{figure*}

Fig.~\ref{fig:hf_energy}(a) illustrates that the input representation $U$, the filtered representation $S$, and the final denoised representation $U_d$ exhibit a distinct three-stage pattern in the proportion of high-frequency energy. Compared with $U$, $S$ exhibits a markedly reduced high-frequency energy ratio, verifying that the multiscale smoothing branch effectively extracts low-frequency trends while suppressing local high-frequency fluctuations. Meanwhile, $U_d$ generally yields a lower high-frequency proportion than $U$ but a higher proportion than $S$. This finding suggests that the module does not directly substitute raw inputs with heavily smoothed results; instead, it leverages a gated residual mechanism to modulate the suppression strength of high-frequency disturbances, striking a balance between denoising efficacy and local information retention.

Fig.~\ref{fig:hf_energy}(b) presents the relative reduction ratios for categories $U\rightarrow S$ and $U\rightarrow U_d$. Compared with the strong low-pass filtering effect of category $U\rightarrow S$, category $U\rightarrow U_d$ exhibits a milder reduction magnitude. For prediction horizons of 96 and 192, the relative reduction ratios of $U\rightarrow U_d$ reach $33.3\%$ and $28.5\%$ on ETTh1, $35.1\%$ and $35.2\%$ on ETTh2, and $62.8\%$ and $61.6\%$ on Weather, respectively. These results further verify that the final denoised representation, $U_d$, is not a straightforward smoothed output; instead, it selectively suppresses high-frequency residuals while retaining the primary structure of the original time series and informative local variations. Specifically, this design mitigates the risk of short-term disturbances propagating into the cross-period representation space before periodic rearrangement, while avoiding critical information loss caused by over-smoothing.
\FloatBarrier

\subsection{Efficiency Analysis}
To quantify the computational efficiency of the proposed model, this work conducts efficiency experiments on the Traffic dataset. All compared models are evaluated under identical experimental settings: the look-back window and prediction horizon are both set to 720, the hidden dimension is 128, and the batch size is 4. All efficiency metrics, including parameter count, multiply-accumulate (MAC) operations, peak GPU memory usage, and per-iteration runtime, are measured under the same hardware conditions.

\begin{table}[t]
\centering
\caption{Efficiency comparison on the Traffic dataset. The look-back length and forecast horizon are both set to 720, the hidden size is fixed to 128, and the batch size is set to 4. The reported metrics include parameters, MACs, peak GPU memory, and epoch time.}
\label{tab:efficiency}
\papertablestyle
\setlength{\tabcolsep}{1pt}
\begin{tabular*}{\columnwidth}{@{\extracolsep{\fill}}lllll@{}}
\toprule
Model & Params & MACs & Memory & Time \\
\midrule
PatchTST \cite{ref5}        & 8.71 M    & 38.05 G   & 16149.2 MB & 662.3 s \\
iTransformer \cite{ref6}  & 715.34 K & 617.65 M & 1141.2 MB  & 114.2 s \\
FITS \cite{ref4}               & 10.51 K   & 8.94 M     & 211.8 MB    & 52.1 s  \\
DLinear \cite{ref3}         & 1.04 M    & 893.72 M & 154.0 MB    & 41.7 s  \\
SimpleTM  \cite{ref11}  & 291.90 K & 414.42 M & 226.3 MB    & 114.7 s \\
TQNet \cite{ref9}          & 2.44 M    & 187.13 M & 327.6 MB    & 31.3 s \\
\midrule
SparseTSF/MLP \cite{ref10}  & 7.86 K   & 174.40 M & 204.0 MB   & 52.6 s  \\
Ours                                     & 18.95 K  & 246.81 M & 600.7 MB   & 64.8 s  \\
\bottomrule
\end{tabular*}
\end{table}

As presented in Table~\ref{tab:efficiency}, the proposed model does not achieve optimal values in parameter count or runtime. Compared with SparseTSF/MLP \cite{ref10}, the model's parameter count increases by 11K, and per-iteration runtime rises from 52.6 s to 64.8 s. The result indicates that the newly introduced modules incur only minor computational overhead, while the overall computational scale remains low. With only 18.95K parameters, the model is substantially more lightweight than Transformer-based architectures such as PatchTST \cite{ref5} and iTransformer \cite{ref6}. Compared with other lightweight baselines, the proposed model achieves superior forecasting accuracy with a limited parameter budget, striking an effective balance between prediction performance and computational efficiency. In summary, the model's performance improvements are accompanied by only marginal computational overhead, and its overall lightweight characteristics are well preserved.

\subsection{Sensitivity Analysis of the Look-back Window}
\begin{figure*}[b]
\centering
\includegraphics[width=0.95\textwidth]{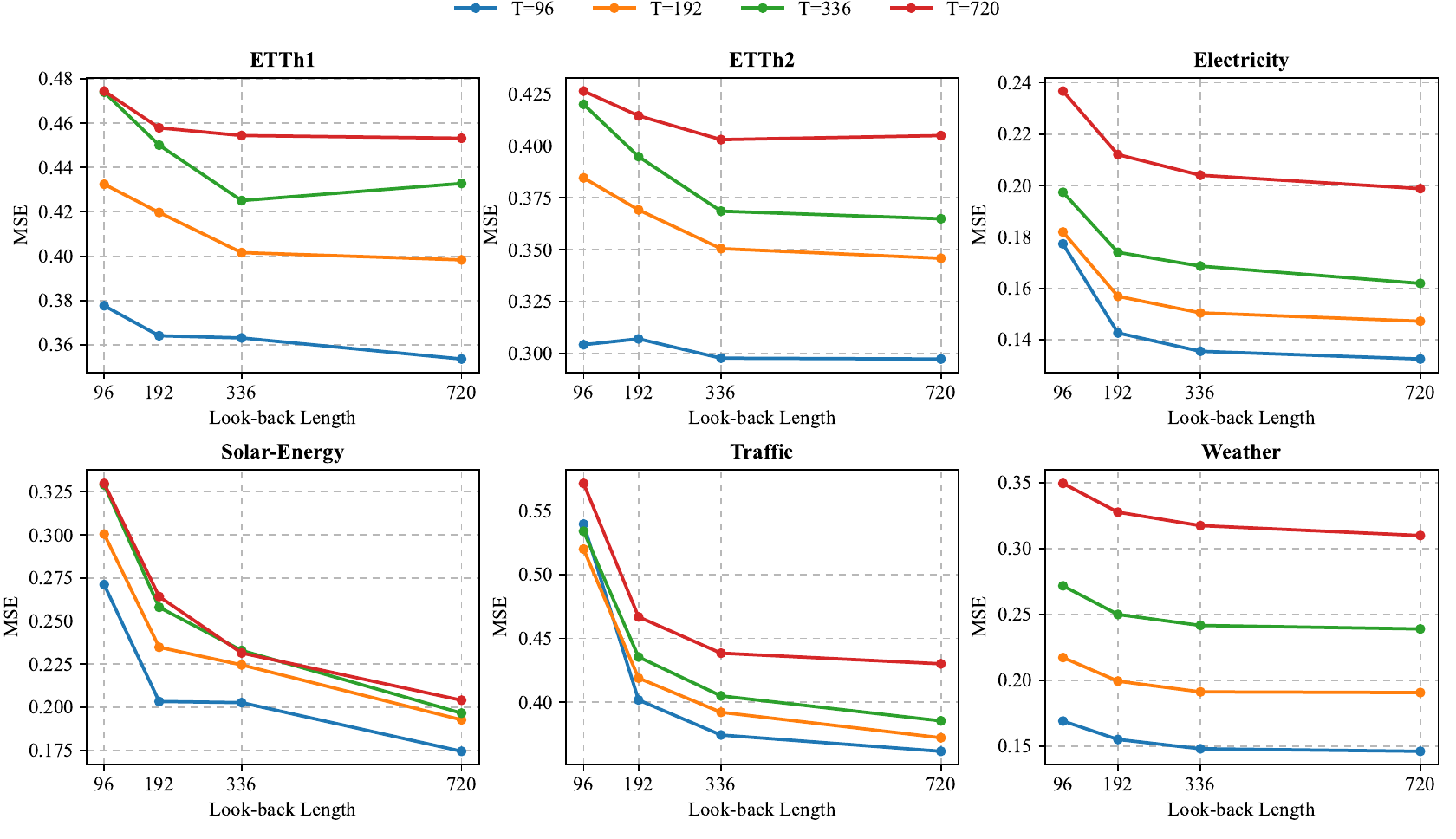}
\par\smallskip
\refstepcounter{figure}\label{fig:lookback}
\makebox[\textwidth][l]{%
\begin{minipage}{\textwidth}
\small\textbf{Fig.~\thefigure.}\enspace
Sensitivity analysis of look-back length. MSE results of TA-SparseMG on six benchmark datasets (ETTh1, ETTh2,
Electricity, Solar-Energy, Traffic, Weather) under varying look-back lengths. The x-axis represents the look-back length, the y-axis denotes the MSE, and each curve corresponds to a specific prediction horizon $T$. Overall, as the look-back length increases from 96 to 720, MSE generally declines across most datasets and prediction horizons, indicating that TA-SparseMG effectively leverages longer historical contexts, exhibits strong long-context modeling capacity, and is robust to input length.
\end{minipage}}
\end{figure*}

To evaluate the model's adaptability to varying look-back window lengths, this work conducts comparative experiments on six benchmark datasets. The look-back window length is $L\in\{96,192,336,720\}$, and the prediction horizon is $T\in\{96,192,336,720\}$. All model architectures and training hyperparameters remain fixed; only the input time series length is adjusted to assess the model's ability to leverage long-term historical context. For LTSF, longer input windows encode more periodic features and temporal evolution patterns, but they also tend to introduce redundant information, local perturbations, and nonstationary components. Whether model performance improves consistently with increasing input length serves as a reliable indicator of its LTS modeling capacity and input adaptability.

As shown in Fig.~\ref{fig:lookback}, the model's prediction error generally decreases as the look-back window extends across most datasets and prediction horizons. The model effectively extracts information from long-term historical contexts without performance degradation due to time-series redundancy or noise. Increasing the look-back window from 96 to 720 reduces the average MSE from 0.354 to 0.287, corresponding to an 18.8\% relative reduction, confirming that long-term observational information can be reliably translated into forecasting gains. The model maintains strong forecasting performance with short input time series while effectively capturing long-range dependencies, demonstrating excellent input-length adaptability and robust LTS modeling.
\FloatBarrier

\section{Conclusions and Outlook}
\label{sec:conclusion}
To address the inherent trade-off among forecasting accuracy, distributional adaptability, and robustness to noise in lightweight LTSF frameworks, this work presents TA-SparseMG, an enhanced lightweight cross-period forecasting model. Built upon SparseTSF's cross-period sparse backbone, it integrates three dedicated components: trend-aware reversible instance normalization, scale-adaptive gated denoising, and a multiscale gated attention predictor to sequentially perform distribution alignment, feature purification, and cross-period mapping optimization. Extensive experiments across six mainstream LTSF benchmarks with diverse prediction horizons demonstrate that TA-SparseMG consistently achieves superior, stable performance. Ablation studies validate the individual efficacy of each module: the denoising component effectively mitigates high-frequency perturbations, and the multiscale attention predictor strengthens the model's adaptability to complex temporal patterns. Overall, TA-SparseMG demonstrates that optimizing the distribution recovery, feature processing, and prediction modules while constraining model complexity can effectively improve the overall performance and robustness of lightweight cross-period forecasting models.

Nevertheless, several limitations remain in the current framework. First, forecasting performance is sensitive to hyperparameters such as the period length, convolution kernel size, and gating coefficients. Optimal configurations vary across downstream tasks, and automatic parameter adaptation still requires further improvement. Second, TA-RevIN performs distribution extrapolation based on local window statistics. Despite its structural simplicity, it exhibits limited capacity to handle abrupt regime shifts, strong nonlinearity, and multi-stage distributional drift. Third, the prediction module is shared across all variable dimensions in the current architecture, which insufficiently exploits inter-variable correlations and thus yields marginal performance gains on high-dimensional, strongly coupled datasets.

Future research can be extended in three directions. First, data-driven adaptive strategies for period selection and denoising scale adjustment can be developed to reduce reliance on manual hyperparameter tuning. Second, the statistical estimation module can be further optimized to enhance the model's robustness to complex nonstationary time series with abrupt changes. Third, lightweight cross-variable interaction mechanisms can be introduced to improve model adaptability in high-dimensional forecasting scenarios.

\bibliographystyle{elsarticle-num-names}
\bibliography{Manuscript_refs}
\end{document}